\documentclass{article}

\usepackage[final,nonatbib]{ml4h_2018}

\usepackage[utf8]{inputenc} 
\usepackage[T1]{fontenc}    
\usepackage{hyperref}       
\usepackage{url}            
\usepackage{booktabs}       
\usepackage{amsfonts}       
\usepackage{nicefrac}       
\usepackage{microtype}      

\usepackage{comment}
\usepackage{graphicx}
\usepackage{float}
\usepackage{subcaption}
\usepackage[titletoc,title,page]{appendix}

\usepackage{lscape}
\usepackage{booktabs}

\title{Generalizability of predictive models for intensive care unit patients}

\author{
  Alistair E. W. Johnson, Tom J. Pollard\\
  Institute of Medical Engineering \& Science\\
  Massachussetts Institute of Technology\\
  Cambridge, MA, USA\\
  \texttt{\{aewj, tpollard\}@mit.edu}\\
  \And
  Tristan Naumann\\
  Microsoft Research\\
  Redmond, WA, USA\\
  \texttt{tristan@microsoft.com}\\
}

\begin{document}

\maketitle

\begin{abstract}
A large volume of research has considered the creation of predictive models for clinical data; however, much existing literature reports results using only a single source of data. In this work, we evaluate the performance of models trained on the publicly-available eICU Collaborative Research Database. We show that cross-validation using many distinct centers provides a reasonable estimate of model performance in new centers. We further show that a single model trained across centers transfers well to distinct hospitals, even compared to a model retrained using hospital-specific data. Our results motivate the use of multi-center datasets for model development and highlight the need for data sharing among hospitals to maximize model performance.
\end{abstract}

\section{Introduction}

The intensive care unit (ICU) admits severely ill patients in order to provide radical, life-saving treatment. ICUs frequently have high staff to patient ratio in order to facilitate continuous monitoring of all patients and to ensure that deterioration in patient condition is detected, and corrected, before it becomes fatal. This approach is demonstrated to improve patient outcomes~\cite{Kane2007}. As a result, the ICU is a data-rich environment.

Major efforts have been placed in utilizing this data to quantify patient health and to predict future outcomes. One of the most immediately relevant outcomes to the ICU is patient mortality. The APACHE system, first published in \cite{Knaus1981}, provides predictions for patient mortality based upon data collected in the ICU. While the initial system was based off expert rules, later updates used data driven methods~\cite{Zimmerman2008}.
Numerous other prediction systems have also been developed~\cite{Knaus1991,LeGall1983use,LeGall1993,Vincent1996,LeGall1996logistic,Johnson2013oasis}; 
for a review of severity illness scores in the ICU, see \cite{strand2008severity} and \cite{keegan2011severity}.

With recent advances in both machine learning and hardware for data archiving, research has begun to return to building better prediction models using more detailed granular data.
The use of mortality prediction models to evaluate ICUs as a whole has found great success, both for capturing patient severity of illness and for risk-adjustment of patient populations~\cite{hug2009icu,physionet2012,johnson2012patient,citi2012physionet,vairavan2012prediction,macavs2012linear,mcmillan2012icu,lehman2012risk,ghassemi2014unfolding,ghassemi2015multivariate,caballero2015dynamically,luo2016predicting}.

We hypothesize that a model trained on ICU data from one institution will perform considerably worse when assessed at other ICUs, even when those institutions use the same vendor for collection of data. This expectation of model generalizability reflects known distributional differences in data that result from differing care and data-recording practices among ICUs. Such distributional differences complicate model generalizability in the absence of specialized approaches \cite{wiens2014study, gong2015instance}. We further investigate a learning approach for deploying a prediction model, comparing a transferred model to one that is developed using only local data. We hypothesize that the local model will initially have poor performance, but with sufficient sample size will outperform the transferred model.

\section{Data}

We utilize the eICU Collaborative Research Database (eICU-CRD) v2.0~\cite{eicu}, a publicly available multi-center database sourced from the Philips eICU programme, a tele-medicine initiative where healthcare workers remotely monitor acutely ill patients. All data in eICU-CRD has been de-identified.
The eICU-CRD contains 200,859 distinct unit stays, from which we exclude non-ICU stays and ICU stays for which an APACHE IVa score is not available. The last exclusion criteria implicitly applies APACHE IVa exclusion criteria in order to mitigate issues around repeated sampling, ensure sufficient support, and remove admissions which appear in the database for administrative purproses~\cite{Zimmerman2008}.
After applying these exclusions, we removed hospitals which had less than 500 stays, as we aimed to evaluate performance hospital-wise.
The final cohort contained 50,106 ICU stays across 46 hospitals. Details of this cohort can be found in Appendix~\ref{app:cohort}, which contains demographic data generated using \texttt{tableone}~\cite{pollard2018tableone}. 

\paragraph{Feature Extraction}
For each ICU stay, we extracted data from a fixed window of length $W=24$ (hours) starting on ICU admission.
Features extracted from the window of fixed size $W$ are detailed in Appendix~\ref{app:features}. The features were extracted from a number of physiologic and laboratory measurements. Notably, no explicit data regarding treatment was extracted (e.g. use of vasopressors, mechanical ventilation, dialysis, etc). 
In addition to those listed in Appendix~\ref{tab:features}, we extracted gender, age, race, and whether the hospital admission was for an elective surgery (binary covariate). A total of 82 features were extracted.

\section{Methods}

We evaluated logistic regression (LR) models trained using L2 regularization,  created using \texttt{scikit-learn v0.20.0}~\cite{scikit}.
The outcome evaluated was in-hospital mortality. The area under the receiver operator characteristic curve (AUROC) was used to evaluate discrimination of models, while the Standardized Mortality Ratio (SMR)\footnote{Standardized Mortality Ratio (SMR) is the observed proportion of patients who died divided by the expected proportion of patient death (given probabilistic risks of patient mortality).} was used to evaluate the calibration of models.

\paragraph{Experiments}
We split our cohort evenly into training and test sets hospital-wise. We train a single LR model on the training data and evaluate it on each hospital separately via the AUROC and the SMR. We experimented with domain transfer but it did not noticably impact model performance (Appendix~\ref{app:domain_transfer}).

For our second experiment, we select a single hospital and set aside 500 patients. We then iteratively sample sets of 200 patients without replacement as our recalibration dataset, i.e. we increase the size of this recalibration set by 200 patients each iteration.
At all iterations, we (i) recalibrate\footnote{Recalibration involves training a univariable logistic regression with hospital mortality as the outcome, effectively shifting and scaling model predictions.} the previous model developed using all hospitals in the training set (``transferred`` model), and (ii) develop a new model using only data available at the current iteration (``local'' model). We evaluate models on the 500 patients initially held out. To ensure stability of estimates we repeat this procedure 50 times after randomly reshuffling data and subsequently plot average performance for the  transferred and local models.

We recognize that it can be difficult to fully specify cohort definitions and analyses approaches in critical care~\cite{johnson:mlhc17-reproduce}, and as such we have made our code available for closer study \cite{alistair_johnson_2018_1971219}.\footnote{\url{https://github.com/alistairewj/icu-model-transfer}}

\section{Results}

\begin{figure}
 \centering
 \includegraphics[width=0.8\textwidth]{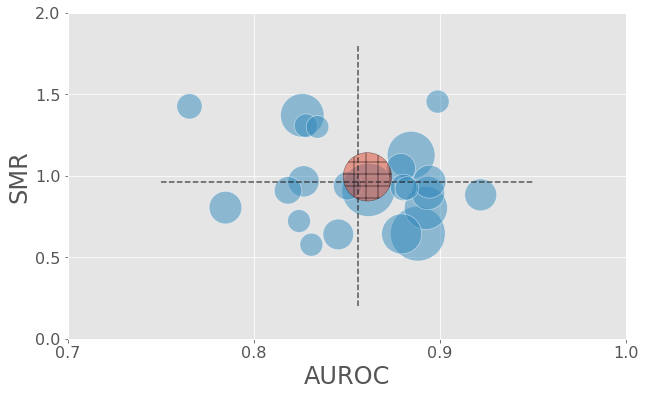}
 \caption{Five fold cross validation (CV) performance of a LR model on the training set (red hatched circle) and for a single model learned from the training set and evaluated on held out hospitals (blue circles). Dashed lines indicate the average of held out performance, and circle size corresponds to hospital size (red hatched circle size is decimated for clarity).}
 \label{fig:exp1}
\end{figure}

The AUROC across 5-fold CV on eICU-CRD was 0.854 and the SMR was 0.998. Figure~\ref{fig:exp1} shows AUROC and SMR on held-out hospitals were distributed around the cross-validation performance. The AUROC ranged from 0.688 to 0.933, and the SMR ranged from 0.560 to 1.43.

\begin{figure}
 \centering
 \includegraphics[width=0.8\textwidth]{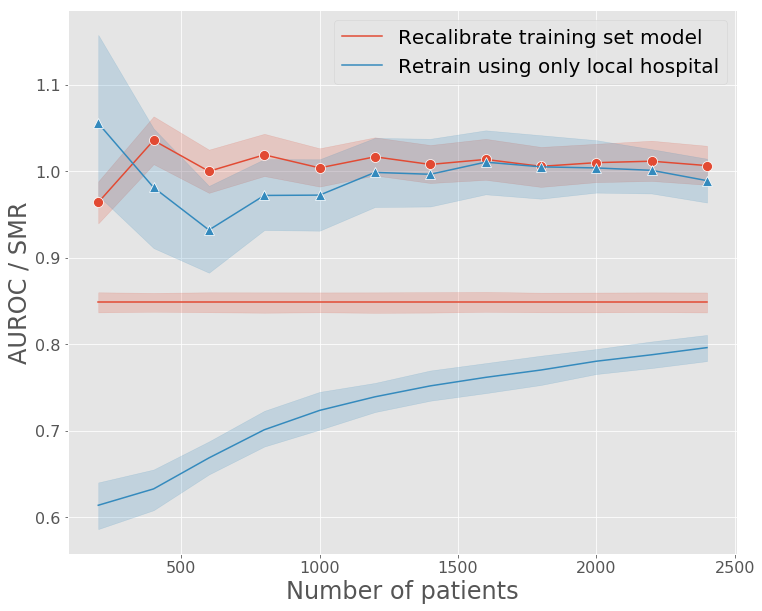}
 \caption{Performance curves when (i) recalibrating a model developed using other hospitals in eICU-CRD (``transferred model''), and (ii) when training a model using only local hospital data (``local model''). Data shown from top left to bottom left: SMRs for the local model (blue triangles), SMRs for the transferred model (red circles), AUROC for the transferred model (red line), and AUROC for the local model (blue line).}
 \label{fig:exp2}
\end{figure}

Figure~\ref{fig:exp2} shows the performance of the recalibration experiment on hospital 73, the largest hospital in the test set. SMRs for both models are reasonable, with the local model initially overpredicting mortality (SMR $>$ 1.0), and eventually converging to the ideal value (1.0). The AUROC of the transferred model is constant at 0.849, which is expected as the recalibration procedure is a monotonic transform which cannot improve pairwise measures such as AUROC. The AUROC of the local model is initially poor (0.614), and after 2,400 patients has significantly improved to 0.796, but is still below the transferred model's AUROC. Other hospitals assessed in this way exhibited similar patterns (two other hospitals are shown in Appendix \ref{app:other_hosp_tfer}).

\section{Discussion and Related Work}

We have evaluated the performance of a model trained using data from many centers and applied to held out hospitals from the continental United States. We found that the cross-validation estimate of performance consistently over-estimated performance on the external hospitals but not by an impractical amount. Notable was the large variance in hospital performance, and there was no consistent pattern of degradation in performance.

Our recalibration experiment assessed how well models transfer to held out hospitals as compared to locally developed models. Local models tended to be well calibrated after reasonable numbers of patients. In terms of discrimination, the local model would likely continue to improve if more data were available in eICU-CRD, but it is noteworthy that even after large sample sizes these models still had inferior discrimination to the transferred models.
An extremely common approach in healthcare is for hospitals to quarantine their data, often citing reasonable privacy concerns, and develop models internally.
While the amount of data archived from healthcare institutions has dramatically increased in the past decade, many prediction problems of interest require specialized cohorts which results in low sample sizes.
Isolating model development to individual hospitals appears to be detrimental to overall model performance, and significant gains are available when data from multiple hospitals is pooled for model development.
Our results indicate that research into safe and secure ways of sharing patient data may be a good avenue for improving performance of predictive models.


There are limitations to our study. We only evaluated mortality prediction models, and did not assess other types of predictive models. However, we believe our results will hold for other predictive models in the ICU. Our feature set was also a small subset of what is available in eICU-CRD. Specifically, we did not utilize any data regarding patient treatment, high granularity vital signs, or patient intake. The incorporation of information from these sources may improve model performance and generalizability. Finally, our results only assess logistic regression as a modelling methodology, though we have found similar results for non-linear approaches such as gradient boosting.

\section{Conclusion}

We have shown that classic in-hospital mortality models trained on publicly-accessible data exhibit varying degrees of decreased performance when applied to external hospital data, both with and without hospital-specific normalization. 
The results agree with common intuition about relative model performance decreasing on external datasets, but provide empirical assessment of the magnitude of this effect.
We further showed that large sample sizes are needed from individual hospitals before performance is close to that achieved by a transfered model from a pooled collection of distinct training data.
This result highlights the importance of data sharing in the development of high performance predictive models.

\section*{Acknowledgements}
This research was funded in part by grants from the National Institutes of Health (NIH): NIH-R01-EB017205, NIH-R01-EB001659, and NIH-R01-GM104987.

\bibliographystyle{plain}
\bibliography{references}

\newpage
\begin{appendices}

\section{Demographics for eICU-CRD}
\label{app:cohort}

Demographics for the cohort are shown in Table \ref{tab:demographics}.

\begin{table}[htbp]
\centering
\begin{tabular}{ll}
\toprule
Demographic & Value \\
\midrule
Number of ICU stays &             50106 \\
Hospital LOS (days) &  5.53 [3.27,9.13] \\
Male &     27833 (55.55) \\
Female   &     22273 (44.45) \\
Age (years)   &   63.00 [52.00,74.00] \\
Elective Surgery  &     11348 (22.65) \\
Race \\
\ White  &     40998 (81.82) \\
\ African Americans &      6720 (13.41) \\
\ Asian &        631 (1.26) \\
\ Hispanic &       1757 (3.51) \\
\\
\end{tabular}
\caption{Demographics of the eICU-CRD cohort. Binary variables are represented as a count with percentage, normally distributed continuous variables are represented as means with standard deviations, and non-normally distributed variables are represented as medians with 25th and 75th percentiles.}
\label{tab:demographics}
\end{table}

\section{Full list of features and window sizes}
\label{app:features}

\begin{table}[htbp]
\centering
\begin{tabular}{c|p{2.7cm}|p{7.0cm}}
Time window & Feature \mbox{extracted} & Variables \\ \hline
$[t_{i,w} - W, t_{i,w}]$ & First, Last & Heart rate, Systolic/Diastolic/Mean blood pressure, Respiratory rate, Temperature, Oxygen Saturation, Glasgow coma scale \\
&& \\
$[t_{i,w} - W - 24, t_{i,w}]$ & \* First, last & Base excess, Calcium, Partial pressure of oxygen in arterial blood, Partial pressure of carbon dioxide in arterial blood, pH, Ratio of partial pressure of oxygen to fraction of oxygen inspired, Total carbon dioxide concentration \\
&& \\
$[t_{i,w} - W - 24, t_{i,w}]$ & \* First, last & Albumin, Blood urea nitrogen, Immature band forms, Bicarbonate, Bilirubin, Creatinine, Hematocrit, Hemoglobin, Lactate, Platelet, Potassium, International Normalized Ratio, Sodium, White blood cell count \\
&& \\
$[t_{i,w} - W, t_{i,w}]$ & Sum & Urine output \\
\\
\end{tabular}
\caption{Features extracted during the first day of the patient's ICU stay. Blood gases and laboratory measurements have the same feature extraction (first, last), but are separated for clarity. \* Note that some features have had their window extended backward an extra 24 hours due to infrequent collection.}
\label{tab:features}
\end{table}

Features extracted used consistent functional forms: the first, last, or sum (in the case of urine output) value was extracted from all measurements of the variable made within first 24 hours. As laboratory values are less frequently sampled, this window was extended 24 hours backward for these measurements.

\section{Domain transfer for improving model performance}
\label{app:domain_transfer}

\begin{figure}[ht]
 \centering
 \includegraphics[width=0.8\textwidth]{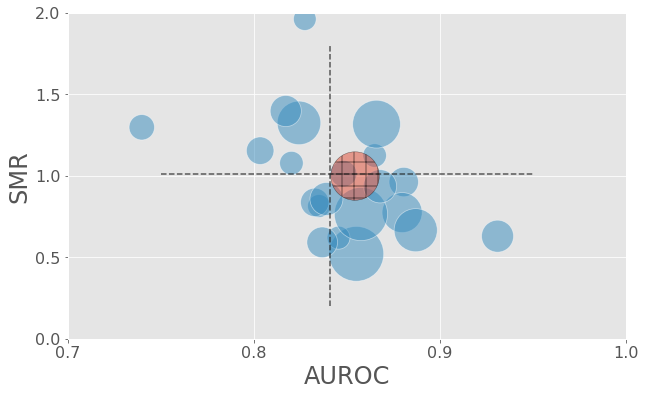}
 \caption{Evaluation of domain transfer on eICU-CRD. Five fold cross validation (CV) performance of a LR model on the training set (red hatched circle) and for a single model learned from the training set and evaluated on held out hospitals (blue circles). Held out hospital data was rescaled using their respective means and standard deviations. Dashed lines indicate the average of held out performance, and circle size corresponds to hospital size (red hatched circle size is decimated for clarity).}
 \label{fig:domain_transfer}
\end{figure}

For each hospital in the test set of eICU-CRD, we rescaled the data using means and standard deviations as calculated on the test set hospital. We compare this to our standard approach which uses the mean and standard deviation of the entire training set to rescale the data. Figure \ref{fig:domain_transfer} shows the performance across test hospitals as measured by the AUROC and SMR. We found no noticably difference between the two methods.

\section{Transfer performance in other hospitals}
\label{app:other_hosp_tfer}

\begin{figure}[ht]
 \centering
 \includegraphics[width=0.8\textwidth]{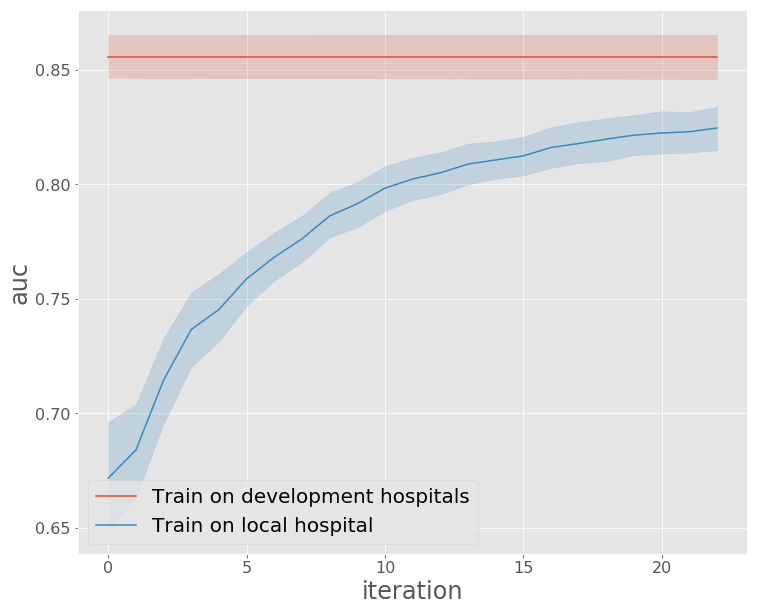}
 \caption{Performance curve when (i) recalibrating a model developed using other hospitals in eICU-CRD (``transferred model''), and (ii) when training a model using only local hospital data (``local model''). The top red line shows the AUROC for the transferred model, and the lower blue line shows the AUROC for the local model.}
 \label{fig:tfer1}
\end{figure}

\begin{figure}[ht]
 \centering
 \includegraphics[width=0.8\textwidth]{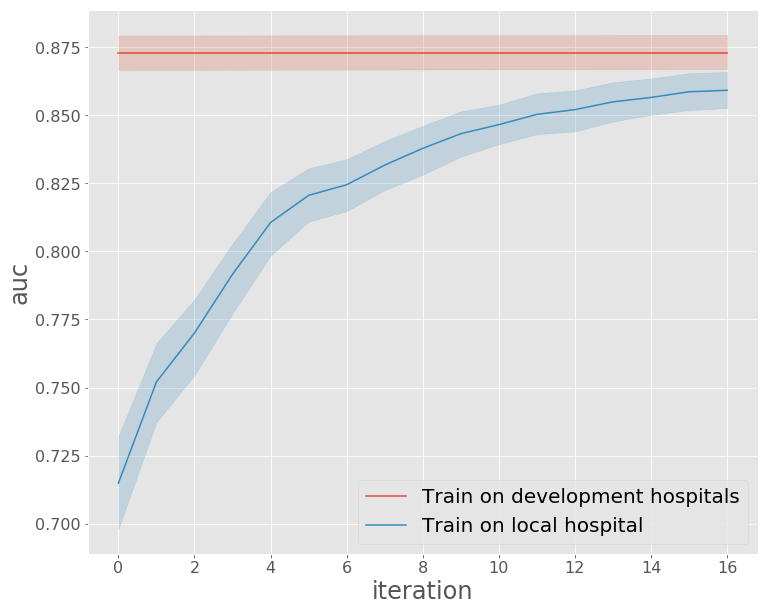}
 \caption{Performance curve when (i) recalibrating a model developed using other hospitals in eICU-CRD (``transferred model''), and (ii) when training a model using only local hospital data (``local model''). The top red line shows the AUROC for the transferred model, and the lower blue line shows the AUROC for the local model.}
 \label{fig:tfer2}
\end{figure}

Figure \ref{fig:tfer1} shows AUROC performance of a model developed on the eICU-CRD training set and recalibrated iteratively to a held-out hospital dataset as described in the paper. Figure \ref{fig:tfer2} shows the same results for a distinct hospital.

\end{appendices}
\end{document}